# Hierarchical Web Page Classification Based on a Topic Model and Neighboring Pages Integration

Wongkot Sriurai
Department of Information Technology
King Mongkut's University of Technology North Bangkok,
Bangkok, Thailand

Phayung Meesad
Department of Teacher Training in Electrical Engineering, King Mongkut's University of Technology North Bangkok,
Bangkok, Thailand

Choochart Haruechaiyasak
Human Language Technology Laboratory
National Electronics and Computer Technology Center (NECTEC),
Bangkok, Thailand

*Abstract*— **Most Web page classification models typically apply the bag of words (BOW) model to represent the feature space. The original BOW representation, however, is unable to recognize semantic relationships between terms. One possible solution is to apply the topic model approach based on the Latent Dirichlet Allocation algorithm to cluster the term features into a set of latent topics. Terms assigned into the same topic are semantically related. In this paper, we propose a novel hierarchical classification method based on a topic model and by integrating additional term features from neighboring pages. Our hierarchical classification method consists of two phases: (1) feature representation by using a topic model and integrating neighboring pages, and (2) hierarchical Support Vector Machines (SVM) classification model constructed from a confusion matrix. From the experimental results, the approach of using the proposed hierarchical SVM model by integrating current page with neighboring pages via the topic model yielded the best performance with the accuracy equal to 90.33% and the F1 measure of 90.14%; an improvement of 5.12% and 5.13% over the original SVM model, respectively.**

## I. INTRODUCTION

Due to the rapid growth of Web documents (e.g., Web pages, blogs, emails) on the World Wide Web (WWW), Web page classification has become one of the key techniques for managing and organizing those documents, e.g., as document filtering in information retrieval. Generally, Web page classification applies the technique of text categorization, which uses the supervised machine learning algorithms for learning the classification model [1, 2]. Most previous works on Web page classification typically applied the bag of words (BOW) model to represent the feature space. Under the BOW model, a Web page is represented by a vector in which each dimension contains a weight value (e.g., frequency) of a word (or term) occurring in the page. The original BOW representation, however, is unable to recognize synonyms from a given word set. As a result, the performance of a classification model using the BOW model could become deteriorated.

In this paper, we apply a topic model to represent the feature space for learning the Web page classification model. Under the topic model concept, words (or terms), which are statistically dependent, are clustered into the same topics. Given a set of documents *D* consisting of a set of terms (or words) *W*, a topic model generates a set of latent topics *T* based on a statistical inference on the term set *W*. In this paper, we apply the Latent Dirichlet Allocation (LDA) [3] algorithm to generate a probabilistic topic model from a Web page collection. A topic model can help capture the hypernyms, hyponyms and synonyms of a given word. For example, the words "vehicle" (hypernym) and "automobile" (hyponym) would be clustered into the same topic. In addition, the words "film" (synonym) and "movie" (synonym) would also be clustered into the same topic. The topic model helps improve the performance of a classification model by (1) reducing the number of feature dimensions and (2) mapping the semantically related terms into the same feature dimension.

In addition to the concept of topic model, our proposed method also integrates some additional term features from neighboring pages (i.e., parent, child and sibling pages). Using some additional terms from neighboring pages could help increase more evidence for learning the classification model [4, 5]. We used the Support Vector Machines (SVM) [6, 7] as the classification algorithm. SVM has been successfully applied to text categorization tasks [6, 7, 8, 9]. SVM is based on the structural risk minimization principle from computational theory. The algorithm addresses the general problem of learning to discriminate between positive and negative members of a given class of n-dimensional vectors. Indeed, the SVM classifier is designed to solve only the binary classification problem [7]. In order to manage the multi-class classification problem, many researches have proposed hierarchical classification methods for solving the multi-class problem. For example, Dumais and Chen proposed the hierarchical method by using SVM classifier for classifying a large, heterogeneous collection of web content. The study showed that the hierarchical method has a better performance than the flat method [10]. Cai and Hofmann proposed a hierarchical classification method that generalizes SVM based on discriminant functions that are structured in a way that mirrors the class hierarchy. The study showed that the hierarchical SVM method has a better performance than the flat SVM method [11].

Most of the related work presented a hierarchical classification method by using different approaches. However in previous works, the bag of words (BOW) model is used to represent the feature space. In this paper, we propose a new hierarchical classification method by using a topic model and integrating neighboring pages. Our hierarchical classification method consists of two phases: (1) feature representation and (2) learning classification model. We evaluated among three different feature representations: (1) applying the simple BOW model on current page, (2) applying the topic model on current page, and (3) integrating the neighboring pages via the topic model. To construct a hierarchical classification model, we use the class relationships obtained from a confusion matrix of the flat SVM classification model. The experimental results showed that by integrating the additional neighboring information via a topic model, the classification performance under the F1 measure was significantly improved over the simple BOW model. In addition, our proposed hierarchical classification method yielded a better performance compared to the SVM classification method.

The rest of this paper is organized as follows. In the next section we provide a brief review of Latent Dirichlet Allocation (LDA). Section 3 presents the proposed framework of hierarchical classification via the topic model and neighboring pages integration. Section 4 presents the experiments with the discussion on the results. In Section 5, we conclude the paper.

## II. A REVIEW OF LATENT DIRICHLET ALLOCATION

Latent Dirichlet Allocation (LDA) has been introduced as a generative probabilistic model for a set of documents [3, 12]. The basic idea behind this approach is that documents are represented as random mixtures over latent topics. Each topic is represented by a probability distribution over the terms. Each article is represented by a probability distribution over the topics. LDA has also been applied for identification of topics in a number of different areas such as classification, collaborative filtering [3] and content-based filtering [13].

Generally, an LDA model can be represented as a probabilistic graphical model as shown in Figure 1 [3]. There are three levels to the LDA representation. The variables $\alpha$ and $\beta$ are the corpus-level parameters, which are assumed to be sampled during the process of generating a corpus. $\alpha$ is the parameter of the uniform Dirichlet prior on the per-document topic distributions. $\beta$ is the parameter of the uniform Dirichlet prior on the per-topic word distribution. $\theta$ is a document-level variable, sampled once per document. Finally, the variables z and w are word-level variables and are sampled once for each word in each document. The variable $N$ is the number of word tokens in a document and variable $M$ is the number of documents.

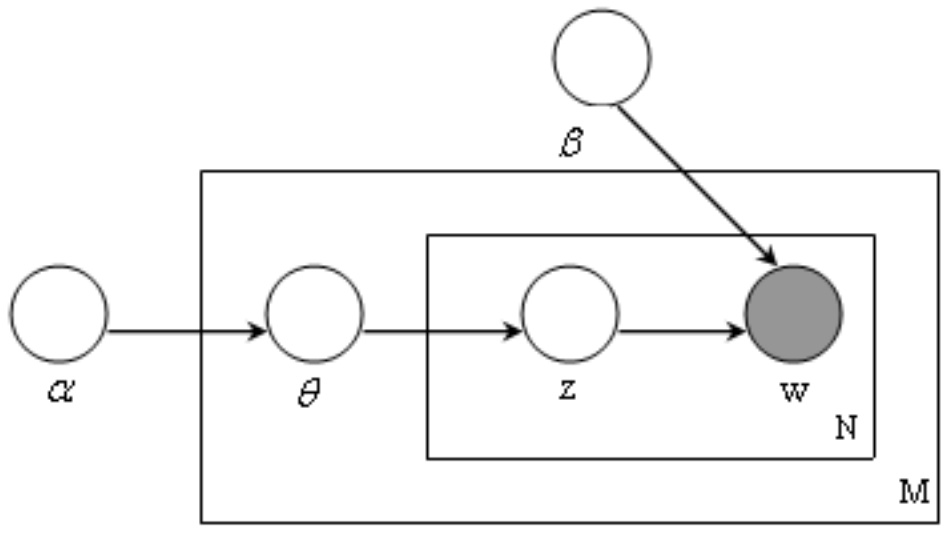

Figure 1. The Latent Dirichlet Allocation (LDA) model

The LDA model [3] introduces a set of $K$ latent variables, called topics. Each word in the document is assumed to be generated by one of the topics. The generative process for each document w can be described as follows:

1. Choose $\theta \sim \text{Dir}(\alpha)$: Choose a latent topics mixture vector $\theta$ from the Dirichlet distribution.
2. For each word $w_n \in W$

   (a) Choose a topic $z_n \sim \text{Multinomial}(\theta)$: Choose a latent topic $z_n$ from the multinomial distribution.

   (b) Choose a word $w_n$ from $P(w_n \mid z_n, \beta)$ a multinomial probability conditioned on the topic $z_n$.

## III. THE PROPOSED HIERARCHICAL CLASSIFICATION FRAMEWORK

Figure 2 illustrates the proposed hierarchical classification framework which consists of two phases: (1) feature representation for learning the Web page classification models, (2) learning classification models based on the Support Vector Machines (SVM). In our proposed framework, we evaluated among three different feature representations: (1) applying the simple BOW model on current page, (2) applying the topic model on current page, and (3) integrating the neighboring pages via the topic model. After the feature representation process, we use the class relationships obtained from a confusion matrix of the flat SVM classification model for building a new hierarchical classification method.

### A. Feature Representation

The process for feature representation can be explained in details as follows.

• **Approach 1 (BOW)**: Given a Web page collection consists of an article collection which is a set of $m$ documents denoted by $D = \{D_0, \ldots, D_{m-1}\}$. In the process of text processing is applied to extract terms. Given a set of terms is represented $W = \{W_0, \ldots, W_{k-1}\}$, where $k$ is the total number of terms. Each term is provided with certain weight $w_i$, which the weight of each term is assigned with term frequency. The set of terms is then filtered by using the feature selection technique, information gain (IG) [1]. Once the term features are obtained, we apply the Support Vector Machines (SVM) to learn the classification model. The model is then used to evaluate the performance of category prediction.

• **Approach 2 (TOPIC_CURRENT):** Given a Web page collection consisting of an article collection which is a set of $m$ documents denoted by $D = \{D_0, \ldots, D_{m-1}\}$. The process of text processing is applied to extract terms. The set of terms is then generated by using the topic model based on the LDA algorithm. The LDA algorithm generates a set of $n$ topics denoted by $T = \{T_0, \ldots, T_{n-1}\}$. Each topic is a probability distribution over $p$ words denoted by $T_i = [w_0^i, \ldots, w_{p-1}^i]$, where $w_j^i$ is a probabilistic value of word $j$ assigned to topic $i$. Based on this topic model, each document can be represented as a probability distribution over the topic set $T$, i.e., $D_i = [t_0^i, \ldots, t_{n-1}^i]$, where $t_j^i$ is a probabilistic value of topic $j$ assigned to document $i$. The output from this step is the topic probability representation for each article. The Support Vector Machines (SVM) is also used to learn the classification model.

• **Approach 3 (TOPIC_ INTEGRATED):** The main difference of this approach from Approach 2 is we integrate the additional term features obtained from the neighboring pages to improve the performance of Web page classification. The process of integrating the neighboring pages is explained as follows.

Figure 3 shows three types of neighboring pages, parent child and sibling pages. Given a Web page (i.e., current page), there are typically incoming links from parent pages, outgoing links to child pages and links from its parent pages to sibling pages. A parent child and sibling pages are collectively referred to as the neighboring pages. Using the additional terms from the neighboring pages could help increase more evidence for learning the classification model.

In this paper, we vary a weight value of neighboring pages from zero to one. A weight value equals to zero means the neighboring pages are not included for the feature representation. Under this approach, terms from different page types (i.e., current, parent, child and sibling) are first transformed into a set of $n$ topics (denoted by $T = \{T_0, \ldots, T_{n-1}\}$) by using the LDA algorithm. The weight values from 0 to 1 are then multiplied to the topic dimension $T_i$ of parent, child and sibling pages. The combined topic feature vector by integrating the neighboring topic vectors with adjusted weight values can be computed by using the algorithm listed in Table 1.

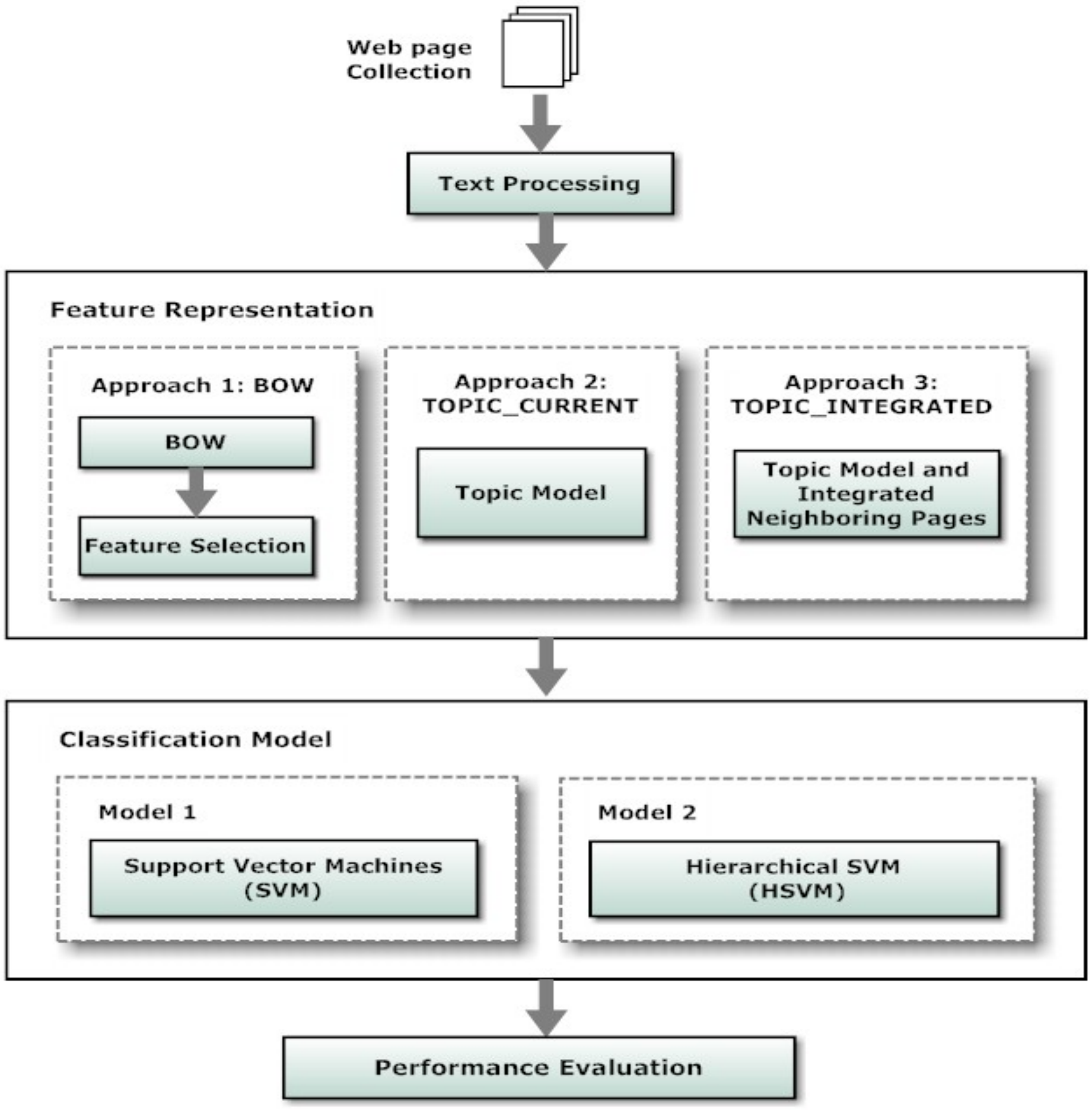

Figure 2. The proposed hierarchical classification framework

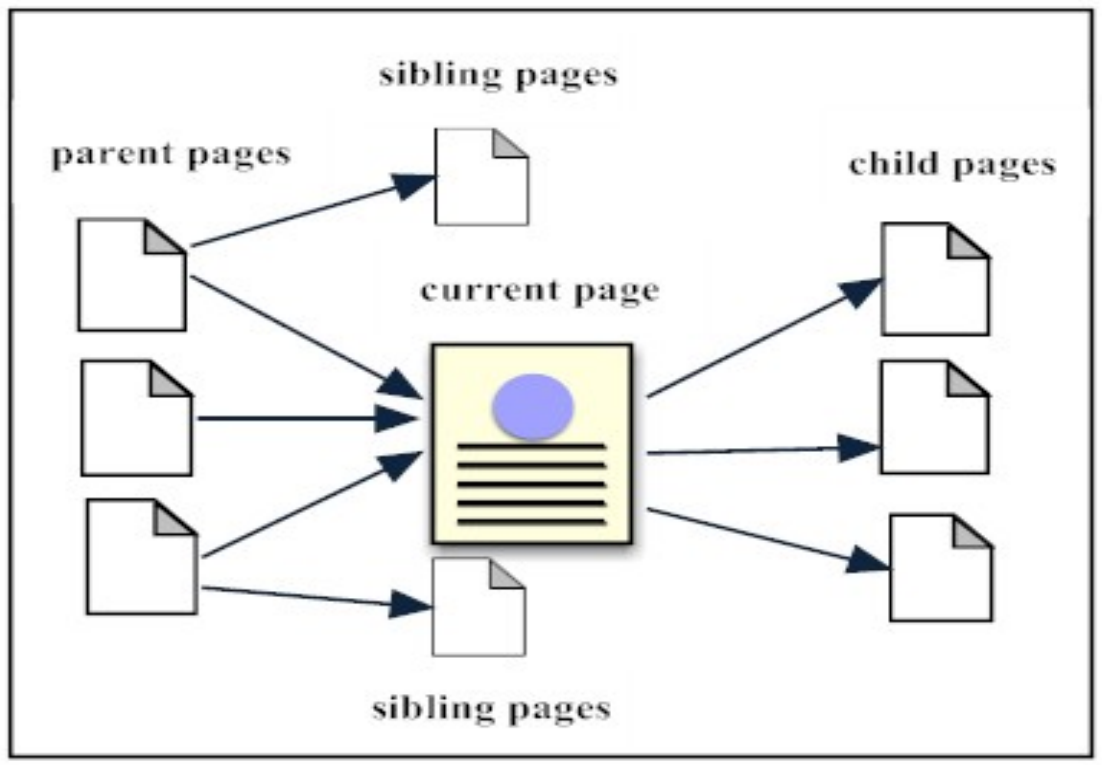

Figure 3. A current Web page with three types of neighboring pages

TABLE I. THE INTEGRATING NEIGHBORING PAGES (INP) ALGORITHM

**Algorithm :** INP

**Input:** CurDT, PDT, CDT, SDT, $W_p$, $W_c$, $W_s$
  **for all** $d_i$ in CurDT **do**
    **for all** $t_j$ in CurDT **do**
      CurDT ← getPValue(CurDT, i, j)
        PP ← getPValue(PDT, i, j) * $W_p$
        PC ← getPValue(CDT, i, j) * $W_c$
        PS ← getPValue(SDT, i, j) * $W_s$
      setPValue(IDT, CurDT + PP + PC + PS, i, j)
    **end for**
  **end for**
**return** IDT

*Parameters and variables*:

- CurDT : document-topic matrix from current page
- PDT : document-topic matrix from parent pages
- CDT : document-topic matrix from child pages
- SDT : document-topic matrix from sibling pages
- IDT : integrated document-topic matrix
- PP : *P*-Value from PDT at specific index
- PC : *P*-value from CDT at specific index
- PS : *P*-value from SDT at specific index
- $W_p$ : weight value for parent pages, $0.0 \leq W_p \leq 1.0$
- $W_c$ : weight value for child pages, $0.0 \leq W_c \leq 1.0$
- $W_s$ : weight value for sibling pages, $0.0 \leq W_s \leq 1.0$
- *P*-value : probability value
- getPValue(*M, r, c*) : function for getting *P*-Value from row *r* and column *c* of matrix *M*
- setPValue(*M, p, r, c*) : function for setting *P*-Value on row *r*, column *c* of matrix *M* with value *p*

The INP algorithm that we present in this paper incorporates term features obtained from the neighboring pages (i.e. parent, child and sibling pages) into the classification model. Using additional terms from the neighboring pages could help increase more evidence for learning the classification model. In this algorithm, we propose a function for varying the weight values of terms from parent pages (PDT), child pages (CDT) and sibling pages (SDT). The probability values from all neighboring pages are integrated with the current page (CurDT) to form a new integrated matrix (IDT).

The process of algorithm begins with the results from the LDA model; that is document-topic matrices from all page types. The algorithm starts by gathering data from document-topic matrices (CurDT, PDT, CDT, SDT) using getPValue function. All P-values of the document-topic matrices are then multiplied by the weight values of each document-topic matrix except for the current page matrix. Finally all *P*-values from four matrices are summed up and then sent to IDT using setPValue function. After the integrating process, we use the IDT matrix for learning the classification model.

### B. Classification Model

Three different feature representation approaches are used as input to classifiers. In this paper, we propose two methods for building the classification models: (1) Model 1: we adopt the SVM to classify feature and (2) Model 2: we presented a new hierarchical classification method by using the class relationships obtained from a confusion matrix for learning a classification model. Each method is described in details as follows.

- **Model 1 (SVM)**: We used the SVM for learning a classification model. The SVM is the machine learning algorithm proposed by Vapnik [7]. The algorithm constructs a maximum margin hyperplane which separates a set of positive examples from a set of negative examples. In the case of examples not linearly separable, SVM uses a kernel functions to map the examples from input space into high dimensional feature space. Using a kernel function can solve the non-linear problem. In our experiments, we used a polynomial kernel. We implemented the SVM classifier by using the WEKA[1] library.

- **Model 2 (HSVM)**: The proposed method is based on SVM classifier, which uses the class relationship obtained from a confusion matrix for building a hierarchical SVM (HSVM). A confusion matrix shows the number of correct and incorrect predictions made by the model compared with the actual classifications of the test data. The size of confusion matrix is *m*-by-*m*, where *m* is the number of classes. Figure 4 shows an example of a confusion matrix from Approach 3 built on a collection of articles obtained from the Wikipedia Selection for Schools. In a confusion matrix, the row corresponds to the actual classes, and the column corresponds to the prediction classes. In this example, for class *art*, the model makes the correct prediction equal to 49 instances and incorrect prediction into class *citizenship* (c) for 1 instance and into class *design and technology* (e) for 5 instances.

[1] Weka. http://www.cs.waikato.ac.nz/ml/weka/

| Class name | | a | b | c | d | e | f | g | h | i | j | k | l | m | n | o |
|---|---|---|---|---|---|---|---|---|---|---|---|---|---|---|---|---|
| Art | a | 49 | 0 | 1 | 0 | 5 | 1 | 0 | 2 | 0 | 1 | 0 | 0 | 2 | 2 | 0 |
| Business Studies | b | 0 | 50 | 11 | 0 | 4 | 3 | 1 | 1 | 3 | 0 | 0 | 0 | 1 | 0 | 0 |
| Citizenship | c | 1 | 4 | 140 | 2 | 1 | 9 | 9 | 8 | 2 | 0 | 0 | 0 | 1 | 1 | 4 |
| Countries | d | 0 | 1 | 0 | 199 | 0 | 0 | 19 | 2 | 0 | 0 | 0 | 0 | 0 | 0 | 0 |
| Design and Technology | e | 4 | 2 | 0 | 0 | 162 | 2 | 8 | 4 | 1 | 0 | 0 | 0 | 2 | 0 | 6 |
| Everyday life | f | 0 | 0 | 6 | 0 | 1 | 254 | 11 | 2 | 0 | 5 | 0 | 2 | 8 | 1 | 24 |
| Geography | g | 0 | 0 | 9 | 10 | 4 | 5 | 681 | 8 | 0 | 0 | 0 | 0 | 1 | 0 | 18 |
| History | h | 0 | 3 | 11 | 0 | 5 | 3 | 14 | 206 | 0 | 3 | 0 | 0 | 25 | 9 | 6 |
| IT | i | 0 | 1 | 0 | 1 | 2 | 1 | 0 | 0 | 63 | 0 | 1 | 0 | 0 | 0 | 0 |
| Language and literature | j | 1 | 0 | 2 | 0 | 3 | 10 | 0 | 7 | 0 | 105 | 0 | 5 | 10 | 3 | 0 |
| Mathematics | k | 0 | 0 | 0 | 0 | 0 | 2 | 0 | 0 | 1 | 0 | 40 | 0 | 0 | 0 | 0 |
| Music | l | 0 | 0 | 0 | 0 | 2 | 2 | 1 | 1 | 0 | 1 | 0 | 113 | 2 | 0 | 0 |
| People | m | 0 | 0 | 1 | 0 | 3 | 3 | 4 | 13 | 1 | 6 | 0 | 2 | 436 | 4 | 2 |
| Religion | n | 0 | 1 | 6 | 0 | 2 | 2 | 2 | 16 | 0 | 4 | 0 | 0 | 1 | 87 | 0 |
| Science | o | 1 | 0 | 0 | 0 | 4 | 19 | 12 | 1 | 0 | 0 | 1 | 0 | 3 | 1 | 828 |

Figure 4. A confusion matrix of Wikipedia Selection for Schools

We used the confusion matrix for constructing a hierarchical structure. First, we need to transform the confusion matrix into a new symmetric matrix, called average pairwise confusion matrix (APCM) by computing average values of pairwise relationships between classes in a confusion matrix (CM). The process of transforming CM into APCM can be explained as follows. Given a confusion matrix CM = [$v_{a,p}$], where *a* denotes each row corresponding to actual classes and *p* denotes each column corresponding to the prediction classes. For the correct prediction, i.e., *a* equals to *p* in CM, we set the value equal to 0 in APCM. If *a* is not equal to *p*, i.e., incorrect prediction, we compute an average value of $v_{a,p}$ and $v_{p,a}$ for a pairwise confusion value at this position. We applied this calculation method for every row and column. For example, in Figure 4, $v_{0,0}$ = 49, *a* is equal to *p* (a correct prediction), $v_{0,0}$ is set equal to 0 in APCM. For $v_{0,2}$ = 1, where *a* = 0, *p* = 2 (*a* is not equal to *p*), an average pairwise confusion value of $v_{0,2}$ and $v_{2,0}$ is equal to 1. The final result of an average pairwise confusion matrix computation is shown in Figure 5. The computation of an average pairwise value is summarized by the following equation:

$$w_{a,p} = \frac{\left(v_{a,p} + v_{p,a}\right)}{2}, \text{ if } a \neq p \quad (1)$$

$$w_{a,p} = 0 \quad , \text{ if } a = p \quad (2)$$

where $w_{a,p}$ = A value from an average pairwise confusion matrix (APCM) at row *a* and column *p*

$v_{a,p}$ = A value from a confusion matrix (CM) at row *a* and column *p*

Once the average pairwise confusion matrix (APCM) is obtained, we construct a dendrogram based on the single link algorithm of hierarchical agglomerative clustering (HAC) [14,15]. Single link clustering is known to be confused by nearby overlapping clusters which merge two clusters with the smallest minimum pairwise distance [14]. To construct our hierarchical classification structure, we adopt the single link algorithm to merge two clusters by selecting maximum average pairwise value in a confusion matrix. We first select a pair of classes which has the maximum average pairwise value in APCM to a dendrogram and select the next highest average pairwise value and go on with this process until all classes are selected into the dendrogram. The final result of dendrogram is shown in Figure 6. For example, an average pairwise value between class *f* and *o* is 21.5, the highest value in APCM, therefore class *f* and class *o* are selected as the first pair in the dendrogram. The second highest value is 19, this value is an average pairwise value between class *h* and *m*, therefore class *h* and class *m* are selected as the second pair. The third highest value is 15 between class *g* and *o*. However, class *o* is already paired with class *f*. Therefore, we take only class *g* to combine with class *f* and class *g* nodes. We perform this process for all remaining classes. Finally, we obtain a complete dendrogram for constructing the hierarchical classification model. The hierarchical classification models are constructed from bottom-up level. With this hierarchical classification structure, classes with lower confusion values are classified before classes with higher confusion. The hierarchical classification model could help improve the performance of multi-class classification method.

| Class name | | a | b | c | d | e | f | g | h | i | j | k | l | m | n | o |
|---|---|---|---|---|---|---|---|---|---|---|---|---|---|---|---|---|
| Art | a | 0 | 0 | 1 | 0 | 4.5 | 0.5 | 0 | 1 | 0 | 1 | 0 | 0 | 1 | 1 | 0.5 |
| Business Studies | b | 0 | 0 | 7.5 | 0.5 | 3 | 1.5 | 0.5 | 2 | 2 | 0 | 0 | 0 | 0.5 | 0.5 | 0 |
| Citizenship | c | 1 | 7.5 | 0 | 1 | 0.5 | 7.5 | 9 | 9.5 | 1 | 1 | 0 | 0 | 1 | 3.5 | 2 |
| Countries | d | 0 | 0.5 | 1 | 0 | 0 | 0 | 14.5 | 1 | 0.5 | 0 | 0 | 0 | 0 | 0 | 0 |
| Design and Technology | e | 4.5 | 3 | 0.5 | 0 | 0 | 1.5 | 6 | 4.5 | 1.5 | 1.5 | 0 | 1 | 2.5 | 1 | 5 |
| Everyday life | f | 0.5 | 1.5 | 7.5 | 0 | 1.5 | 0 | 8 | 2.5 | 0.5 | 7.5 | 1 | 2 | 5.5 | 1.5 | 21.5 |
| Geography | g | 0 | 0.5 | 9 | 14.5 | 6 | 8 | 0 | 11 | 0 | 0 | 0 | 0.5 | 2.5 | 1 | 15 |
| History | h | 1 | 2 | 9.5 | 1 | 4.5 | 2.5 | 11 | 0 | 0 | 5 | 0 | 0.5 | 19 | 13 | 3.5 |
| IT | i | 0 | 2 | 1 | 0.5 | 1.5 | 0.5 | 0 | 0 | 0 | 0 | 1 | 0 | 0.5 | 0 | 0 |
| Language and literature | j | 1 | 0 | 1 | 0 | 1.5 | 7.5 | 0 | 5 | 0 | 0 | 0 | 3 | 8 | 3.5 | 0 |
| Mathematics | k | 0 | 0 | 0 | 0 | 0 | 1 | 0 | 0 | 1 | 0 | 0 | 0 | 0 | 0 | 0.5 |
| Music | l | 0 | 0 | 0 | 0 | 1 | 2 | 0.5 | 0.5 | 0 | 3 | 0 | 0 | 2 | 0 | 0 |
| People | m | 1 | 0.5 | 1 | 0 | 2.5 | 5.5 | 2.5 | 19 | 0.5 | 8 | 0 | 2 | 0 | 2.5 | 2.5 |
| Religion | n | 1 | 0.5 | 3.5 | 0 | 1 | 1.5 | 1 | 12.5 | 0 | 3.5 | 0 | 0 | 2.5 | 0 | 0.5 |
| Science | o | 0.5 | 0 | 2 | 0 | 5 | 21.5 | 15 | 3.5 | 0 | 0 | 0.5 | 0 | 2.5 | 0.5 | 0 |

Figure 5. An average pairwise confusion matrix of Wikipedia Selection for Schools

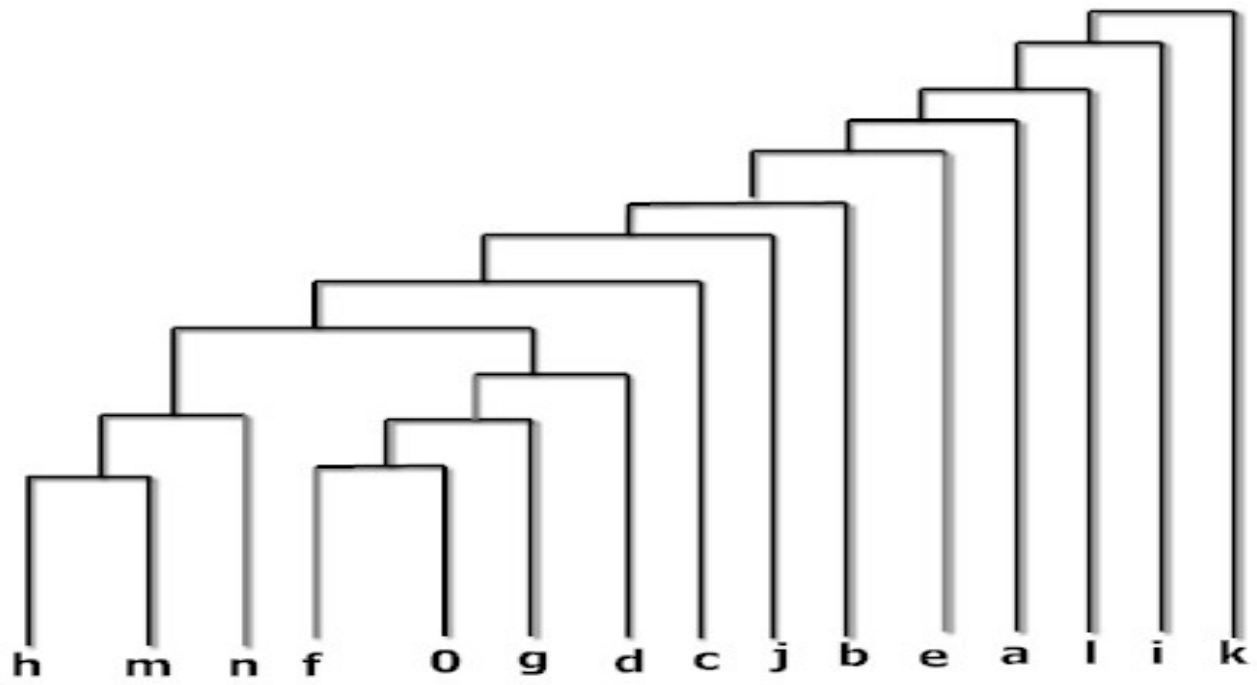

Figure 6. A hierarchies of Wikipedia Selection for Schools

## IV. Experiments and Discussion

### A. Web page collection

In our experiments, we used a collection of articles obtained from the Wikipedia Selection for Schools which is available from the SOS Children's Villages Web site[2]. There are 15 categories: art, business studies, citizenship, countries, design and technology, everyday life, geography, history, IT, language and literature, mathematics, music, people, religion and science. The total number of articles is 4,625.

Table 2 lists the first-level subject categories available from the collection. Organizing articles into the subject category set provides users a convenient way to access the articles on the same subject. Each article contains many hypertext links to other articles which are related to the current article.

TABLE II. The subject categories under the Wikipedia Selection for School

| Category | No. of Articles | Category | No. of Articles |
|---|---|---|---|
| Art | 74 | Business Studies | 88 |
| Citizenship | 224 | Countries | 220 |
| Design and Technology | 250 | Everyday life | 380 |
| Geography | 650 | History | 400 |
| IT | 64 | Language and literature | 196 |
| Mathematics | 45 | Music | 140 |
| People | 680 | Religion | 146 |
| Science | 1068 | | |

### B. Experiments

We used the LDA algorithm provided by the linguistic analysis tool called LingPipe [3] to run our experiments. LingPipe is a suite of Java tools designed to perform linguistic analysis on natural language data. In this experiment, we applyed the LDA algorithm provided under the LingPipe API and set the number of topics equal to 200 and the number of epochs to 2,000.For text classification process, we used WEKA, an open-source machine learning tool, to perform the experiments.

### C. Evaluation Metrics

The standard performance metrics for evaluating the text classification used in the experiments are accuracy, precision, recall and F1 measure [16]. We tested all algorithms by using the *10-fold* cross validation. Accuracy, precision, recall and F1 measure are defined as:

$$Accuracy = \frac{\text{the number of correctly classified test documents}}{\text{total number of test documents}} \quad (3)$$

$$Precision = \frac{\text{the number of correct positive predictions}}{\text{the number of positive predictions}} \quad (4)$$

$$Recall = \frac{\text{the number of correct positive predictions}}{\text{the number of positive data}} \quad (5)$$

$$F1 = \frac{2 \times precision \times recall}{precision + recall} \quad (6)$$

where Accuracy represents the percentage of correct predictions in total predictions. Precision ($P$) is the percentage of the predicted documents for a given category that are classified correctly. Recall ($R$) is the percentage of the documents for a given category that are classified correctly. F1 measure is a single measure that tries to combine precision and recall. F1 measure ranges from 0 to 1 and the higher the better.

### D. Experimental results

We started by evaluating the weight values of neighboring pages under Approach 3. Table 3 shows the results of combination the weight value of neighboring pages on our algorithm. For the SVM model, the best combination of neighboring pages with the accuracy equal to 85.21% and the F1 measure of 0.8501 by weight of parent pages, child pages and sibling pages equal to 0.4, 0.0 and 0.3, respectively and for the HSVM model has the best combination of neighboring pages with the accuracy equal to 90.33% and the F1 measure of 0.9014 by weight the same SVM model. The results showed that using information from parent pages and sibling pages are more effective than child pages for improving the performance of a classification model.

TABLE III. Classification results by integrating neighboring pages

| Models | $W_p$ | $W_c$ | $W_s$ | P | R | F1 | Accuracy (%) |
|---|---|---|---|---|---|---|---|
| SVM | 0.4 | 0.0 | 0.3 | 0.8583 | 0.8337 | 0.8501 | 85.21 |
| HSVM | 0.4 | 0.0 | 0.3 | 0.8984 | 0.9046 | 0.9014 | 90.33 |

From Table 4, the results of classification model based on two models between the SVM model and the hierarchical SVM (HSVM), the approach of integrating current page with the neighboring pages via the topic model (TOPIC_INTEGRATED) yielded a higher accuracy compared to applying the topic model on current page (TOPIC_CURRENT) and applying the BOW model. For the SVM model, on the TOPIC_INTEGRATED approach, the highest accuracy is 85.21%; improvement of 23.96% over the BOW model. For the HSVM model, on the TOPIC_INTEGRATED approach, the highest accuracy is 90.33%; improvement of 4.64% over the BOW model.

[2] SOS Children's Villages Web site. http://www.soschildrensvillages.org.uk/charity-news/wikipedia-for- schools.htm
[3] LingPipe. http://alias-i.com/lingpipe

TABLE IV. EVALUATION RESULTS ON CLASSIFICATION MODELS BY USING THREE FEATURE REPRESENTATION APPROACHES

| Feature Representation Approaches | Classification Models | |
|---|---|---|
| | SVM | HSVM |
| | Accuracy (%) | Accuracy (%) |
| 1. BOW | 61.25 | 85.69 |
| 2. TOPIC_CURRENT | 78.54 | 88.97 |
| 3.TOPIC_INTEGRATED ($W_p$=0.4 ,$W_c$=0.0 & $W_s$ =0.3) | 85.21 | 90.33 |

TABLE V. CLASSIFICATION RESULTS BASE ON THREE FEATURE REPRESENTATION APPROACHES

| Feature Representation Approaches | Classification Models | | | | | |
|---|---|---|---|---|---|---|
| | SVM | | | HSVM | | |
| | P | R | F1 | P | R | F1 |
| 1. BOW | 0.6000 | 0.6610 | **0.6120** | 0.8485 | 0.8541 | **0.8503** |
| 2. TOPIC_CURRENT | 0.7960 | 0.7710 | **0.7840** | 0.8886 | 0.8908 | **0.8891** |
| 3.TOPIC_INTEGRATED ($W_p$=0.4 ,$W_c$=0.0 & $W_s$ =0.3) | 0.8583 | 0.8337 | **0.8501** | 0.8984 | 0.9046 | **0.9014** |

Table 5 shows the experimental results of three feature representation approaches by using two models between the SVM model and the hierarchical SVM (HSVM) model for the learning classification model. From this table, the approach of integrating current page with neighboring pages via the topic model (TOPIC_INTEGRATED) yielded a higher performance compared to applying the topic model on current page (TOPIC_CURRENT) and application of the BOW model. The HSVM classification model yielded a higher performance compared to the SVM classification model in all three feature representation approaches.

The results of classification model based on the SVM model, applying the TOPIC_CURRENT approach helped improve the performance over the BOW by 17.2% based on the F1 measure and applying the TOPIC_INTEGRATED approach, yielded the best performance with the F1 measure of 85.01%; improvement of 23.81% over the BOW model. For the learning classification model based on the HSVM model, applying the TOPIC_CURRENT approach helped improve the performance over the BOW by 3.88% based on the F1 measure and applying the TOPIC_INTEGRATED, yielded the best performance with the F1 measure of 90.14%; improvement of 5.11% over the BOW model. The approach of integrating current page with the neighboring pages via the topic model (TOPIC_INTEGRATED) and using the HSVM model, however, yielded the best performance with the F1 measure of 90.14%; improvement of 5.13% over the TOPIC_INTEGRATED approach by using the SVM model. Thus, integrating the additional neighboring information, especially from the parent pages and sibling pages, via a topic model could significantly improve the performance of a classification model. The reason is due to the parent pages often provide terms, such as in the anchor texts, which provide additional descriptive information of the current page.

## V. CONCLUSIONS

To improve the performance of Web page classification, we proposed a new hierarchical classification method based on a topic model and by integrating the additional term features obtained from the neighboring pages to improve the performance of Web page classification. We applied the topic model approach based on the Latent Dirichlet Allocation algorithm to cluster the term features into a set of latent topics. Terms assigned into the same topic are semantically related. Our hierarchical classification method consists of two phases: (1) feature representation by using a topic model and integrating neighboring pages, and (2) hierarchical Support Vector Machines (SVM) classification model constructed from a confusion matrix. From the experimental results, the approach of integrating current page with the neighboring pages via the topic model yielded a higher performance compared to applying the topic model on current page and applying the BOW model. For learning classification model, the hierarchical SVM classification model yielded a higher performance compared to the SVM classification model in all three feature representation approaches and integrating current page with the neighboring pages via the topic model approach, however, yielded the best performance with the F1 measure of 90.14%; improvement of 5.11% over the BOW model. The approach of integrating current page with the neighboring pages via the topic model and using the hierarchical SVM classification model yielded the best performance with the accuracy equal to 90.33% and the F1 measure of 90.14%; an improvement of 5.12% and 5.13% over the original SVM model, respectively.

AUTHORS PROFILE

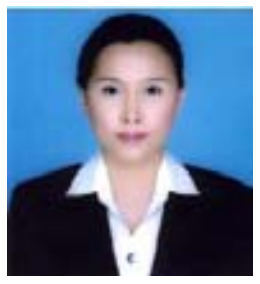

**Wongkot Sriurai** received B.Sc. degree in Computer Science, M.S. degree in Information Technology from Ubon Ratchathani University. Currently, she is a Ph.D. candidate in the Department of Information Technology at King Mongkut's University of Technology North Bangkok. Her current research interests Web Mining, Information filtering and Recommender system.

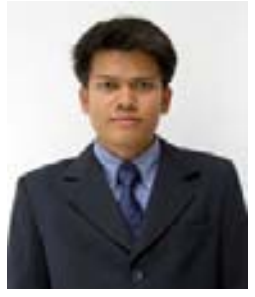

**Phayung Meesad** received the B.S. from King Mongkut's University of Technology North Bangkok M.S. and Ph.D. degree in Electrical Engineering from Oklahoma State University. His current research interests Fuzzy Systems and Neural Networks, Evolutionary Computation and Discrete Control Systems. Currently, he is an Assistant Professor in Department of Teacher Training in Electrical Engineering at King Mongkut's University of Technology North Bangkok Thailand.

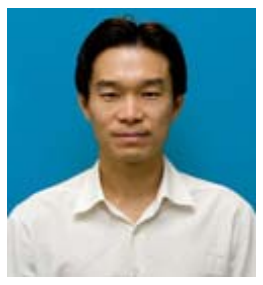

**Choochart Haruechaiyasak** received B.S. from University of Rochester, M.S. from University of Southern California and Ph.D. degree in Computer Engineering from University of Miami. His current research interests Search technology, Data/text/Web mining, Information filtering and Recommender system. Currently, he is chief of the Intelligent Information Infrastructure Section under the Human Language Technology Laboratory (HLT) at National Electronics and Computer Technology Center (NECTEC), Thailand.